\documentclass[11pt,a4paper]{article}
\usepackage[protrusion=true,expansion=true]{microtype}
\usepackage[margin=1in]{geometry}
\usepackage{amsmath,amssymb,amsfonts,amsthm,mathtools}
\usepackage{enumitem}
\usepackage{booktabs}
\usepackage{array}
\usepackage{hyperref}
\usepackage{xcolor}
\usepackage{tikz}
\usepackage{microtype}
\usepackage{setspace}
\usetikzlibrary{arrows.meta,positioning,shapes.geometric,fit,calc}

\usepackage[utf8]{inputenc}
\usepackage[T1]{fontenc}
\usepackage[margin=1in]{geometry}

\usepackage{float}

\usepackage{natbib}

\hypersetup{
    colorlinks=true,
    linkcolor=blue,
    filecolor=magenta,
    urlcolor=cyan,
    citecolor=blue,
}

\hypersetup{colorlinks=true,linkcolor=blue!50!black,citecolor=blue!50!black,urlcolor=blue!50!black}
\setlist[itemize]{topsep=3pt,itemsep=2pt,parsep=0pt}
\setlist[enumerate]{topsep=3pt,itemsep=2pt,parsep=0pt}

\title{When Is Next-Token Prediction Useful?\\
\large Marginalization, Ergodicity, Mixture Identifiability, Local Sufficiency, RAG, Tools, and Programming}
\author{Francesco Corielli}
\date{\today}

\begin{document}
\maketitle

\begin{abstract}
Language models trained on observed sequences are often described as learning the conditional distribution of the next token given previous tokens. This description is useful but incomplete. A model trained on realized token trajectories does not directly observe full conditional laws; it observes samples. Moreover, real language generation is conditioned not only on previous words but also on non-textual circumstances: facts, events, intentions, goals, beliefs, social context, and task-specific constraints. This paper distinguishes three objects that are often conflated: the full conditional language process conditioned on latent circumstances, the marginal text-only process obtained by integrating those circumstances out, and the model-induced distribution learned from finite observed corpora.

Many phenomena discussed here are individually familiar in the literature on language models, including hallucination, grounding failure, retrieval augmentation, tool use, in-context learning, exposure bias, and recursive training on synthetic data. The contribution of the paper is not to claim that these problems are unknown. Rather, it is to place them in a common statistical framework. The central claim is that next-token prediction is epistemically useful only when the observed and augmented context is sufficiently informative about the latent circumstances relevant to continuation. In information-theoretic terms, usefulness requires that the residual conditional mutual information between the next token and omitted circumstances, given the observed context, be small.

The paper applies this framework to heterogeneous corpora, programming, RAG, tools, temperature, synthetic contamination, and prompt-based context injection. RAG and tools are interpreted as conditional-sufficiency devices: they are useful only insofar as retrieved material or external computation makes the remaining latent circumstances irrelevant, or substantially less informative, for continuation. Programming is analyzed as a favorable case because specifications, previous code, tests, error messages, and documentation often textualize much of the relevant latent state.

\end{abstract}

\section{Introduction}

A common statement about language models is that they learn the probability distribution of the next word or token given previous words or tokens, a view descending from the statistical language-modeling tradition initiated by Shannon and developed in modern NLP \citet{shannon1948,shannon1951,manning1999,rosenfeld2000}. This statement captures part of the statistical structure of language modeling, but it conceals several distinct assumptions.

First, language models trained on sequences do not directly observe conditional probability distributions. In maximum-likelihood language modeling, from classical neural language models to contemporary transformers, the training signal is a realized next token and the loss is cross-entropy against that token \cite{bengio2003,mikolov2010,vaswani2017,radford2019,brown2020}. They observe realized trajectories. For each context, they are rewarded for increasing the probability assigned to the observed next token. The conditional law is therefore not an object present in the data; it is an inferred statistical object.

Second, real language generation is not conditioned only on previous words. Human utterances depend on latent and external circumstances: facts, events, communicative intentions, institutional settings, social relations, speaker beliefs, physical reality, and task constraints. This is related to the distinction between linguistic form and grounded meaning emphasized by \cite{bender2020}. A text-only language model has access to these variables only insofar as they are represented, implied, or recoverable from the observed prefix.

Third, even if one defines a marginal text-only conditional distribution by integrating over the missing circumstances, a finite corpus informs that marginal distribution only under strong assumptions. The corpus must be a sufficiently representative realization of a stable process. In stochastic-process language, this requires assumptions analogous to stationarity and ergodicity.

Fourth, even if the model correctly estimates the marginal text-only distribution, this does not guarantee usefulness. The marginal law is useful only when the text prefix is an approximately sufficient statistic for the omitted circumstances relevant to the continuation.

Fifth, real training corpora are heterogeneous mixtures. They contain local regimes in which textual sufficiency holds approximately, and many regimes in which it does not. A model trained on such a mixture may therefore be locally reliable in some domains and unreliable in others, while producing equally fluent outputs everywhere. Mixture models and latent-variable perspectives are standard in statistics \citet{mclachlan2000}; here they are used to clarify when a text prefix identifies a local linguistic regime.

Sixth, heterogeneous training introduces an additional learnability problem. For a model to learn a correct mixture conditional, the textual prefix must provide enough information to infer, at least probabilistically, which component regime generated the prefix. If the regime is identifiable, local conditional laws may be learned. If it is not, the model learns only a blended distribution over heterogeneous continuations.

This paper formalizes these distinctions and derives a criterion for when next-token prediction is useful. It then discusses programming as a favorable case and clarifies the role of Retrieval Augmented Generation (RAG) and tool use as attempts to improve the sufficiency of the conditioning context \cite{lewis2020,borgeaud2022,schick2023,yao2023,karpas2022,gao2023}. The point is deliberately architecture-neutral: the argument applies to any language model trained on observed sequences, from classical statistical language models to contemporary neural models and LLMs.

The paper deliberately does not address a second, even more important, problem of language models. Even if the text-only language distribution were estimated successfully, nothing in that fact guarantees that the generated sequences correspond to true statements. This is a distinct issue. A model may learn a distribution of linguistic continuations and still produce false factual claims, invalid arguments, or wrong arithmetic operations. Standard language models do not contain an intrinsic truth checker or validity checker; their objective is to assign probability to sequences, not to verify correspondence with facts, proofs, executions, or calculations. Consequently, truthfulness depends on two properties that must not be conflated: first, whether the relevant language distribution has been learned correctly; second, whether high-probability continuations in that distribution are true or valid in the target domain. Hallucination and factuality studies make this distinction central, and even in domains such as arithmetic or symbolic reasoning, external execution or verification is often needed to separate plausible generated text from correct results \cite{ji2023,huang2023,kadavath2022,gao2023,schick2023}.

\subsection*{Relation to existing literature and contribution}

The individual ingredients of this paper have many precedents. Prior work has examined grounding and the limits of form-only learning \citet{bender2020}; hallucination and factuality failures \citet{ji2023,huang2023,maynez2020,lin2022,kadavath2022}; retrieval-augmented and retrieval-enhanced language modeling \citet{lewis2020,borgeaud2022}; tool use, program-aided generation, and neuro-symbolic orchestration \citet{karpas2022,yao2023,schick2023,gao2023}; in-context learning and Bayesian or latent-variable interpretations of prompting \citet{brown2020,min2022,xie2022}; exposure bias and decoding effects \citet{he2021,holtzman2020}; and recursive training on synthetic data or model collapse \citet{shumailov2023,shumailov2024,alemohammad2024}.

The contribution of the present paper is not to claim that hallucination, grounding failure, RAG limitations, tool dependence, prompt sensitivity, or recursive contamination are new phenomena. Rather, the contribution is to organize them within a single statistical framework. The central claim is that prediction from a learned text-only marginal law is epistemically useful only when the observed and augmented context is sufficiently informative about the latent circumstances relevant to continuation. RAG, tools, programming contexts, and prompts are therefore not separate exceptions to language modeling; they are different ways of changing the conditioning information available to generation.

\section{Three Different Objects not to be Confused}

Let $X_t$ denote the token at position $t$, and let
\[
X_{\leq t}=(X_1,\ldots,X_t)
\]
be the observed textual history. Let $Z_t$ denote the non-textual circumstances relevant to language production at time $t$. These may include
\[
Z_t=(\text{world state},\text{facts},\text{events},\text{speaker beliefs},\text{goals},\text{intentions},\text{audience},\text{task constraints},\ldots).
\]
It is useful to distinguish three objects.

\subsection{The full conditional language law}

The full conditional law is
\[
p_{\mathrm{full}}(x_{t+1}\mid x_{\leq t},z_t).
\]
This object describes the probability of a continuation given both the textual prefix and the relevant non-textual circumstances. It is the object closest to real language production. A person does not speak only because previous words statistically suggest a continuation. A person speaks in relation to intentions, facts, perceptions, social pressures, tasks, and goals.

\subsection{The marginal text-only conditional law}

If $Z_t$ is unobserved, one can define the marginal text-only conditional distribution:
\[
p_{\mathrm{marg}}(x_{t+1}\mid x_{\leq t})
=
\int p_{\mathrm{full}}(x_{t+1}\mid x_{\leq t},z)p(z\mid x_{\leq t})dz.
\]
This is a theoretical object. It is the conditional distribution of future tokens given past tokens after integrating over latent circumstances. Importantly, this is not the full human language process. It is the full process after relevant circumstances have been averaged out.

\subsection{The model-induced predictive distribution}

A language model with parameters $\theta$ defines
\[
p_\theta(x_{t+1}\mid x_{\leq t}).
\]
At decoding temperature $T$, logits $\ell_i$ are transformed into
\[
p_{\theta,T}(i\mid x_{\leq t})=
\frac{\exp(\ell_i/T)}{\sum_j\exp(\ell_j/T)}.
\]
The model samples from $p_{\theta,T}$, not from $p_{\mathrm{full}}$. Nor can $p_{\theta,T}$ automatically be identified with $p_{\mathrm{marg}}$. The relation between these objects depends on assumptions.

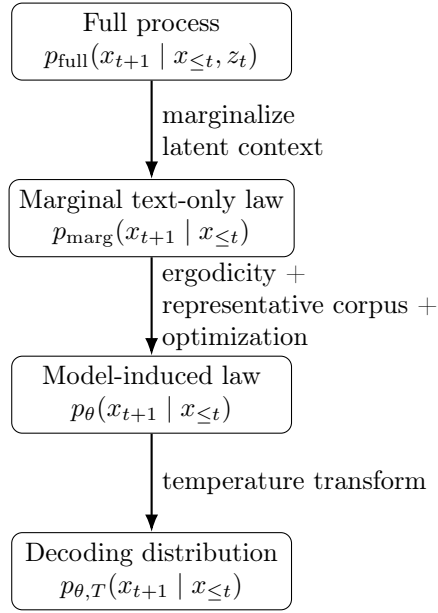
\begin{figure}[ht]
\centering
\begin{tikzpicture}[node distance=1.3cm, every node/.style={font=\small}, box/.style={draw,rounded corners,align=center,minimum width=3.7cm,minimum height=0.9cm}, arr/.style={-Latex,thick}]
\node[box] (full) {Full process\\$p_{\mathrm{full}}(x_{t+1}\mid x_{\leq t},z_t)$};
\node[box,below=of full] (marg) {Marginal text-only law\\$p_{\mathrm{marg}}(x_{t+1}\mid x_{\leq t})$};
\node[box,below=of marg] (model) {Model-induced law\\$p_\theta(x_{t+1}\mid x_{\leq t})$};
\node[box,below=of model] (temp) {Decoding distribution\\$p_{\theta,T}(x_{t+1}\mid x_{\leq t})$};
\draw[arr] (full) -- node[right,align=left]{marginalize\\latent context} (marg);
\draw[arr] (marg) -- node[right,align=left]{ergodicity +\\representative corpus +\\optimization} (model);
\draw[arr] (model) -- node[right]{temperature transform} (temp);
\end{tikzpicture}
\caption{Four distinct distributions. The model samples from the last object. Identifying it with the preceding objects requires assumptions.}
\end{figure}

\section{What Training Actually Observes}

During standard next-token training, the model sees realized token sequences
\[
(x^{(m)}_1,\ldots,x^{(m)}_{T_m}),\qquad m=1,\ldots,M.
\]
For each position $t$, the training signal is the realized pair
\[
(x^{(m)}_{\leq t},x^{(m)}_{t+1}).
\]
The standard maximum-likelihood objective is the cross-entropy loss \cite{bishop2006,cover2006}:
\[
\mathcal{L}(\theta)=
-\sum_{m=1}^{M}\sum_{t=1}^{T_m}\log p_\theta(x^{(m)}_{t+1}\mid x^{(m)}_{\leq t}).
\]
For a single training instance, the loss is minimized by assigning probability one to the observed next token:
\[
p_\theta(x^{(m)}_{t+1}\mid x^{(m)}_{\leq t})=1.
\]
With shared parameters and many related or conflicting contexts, the global optimum cannot generally assign probability one to every observed continuation. The model must compress, interpolate, and generalize.

The immediate training signal is therefore not a full conditional distribution. It is a realized continuation. Thus, the claim that the model learns the next-token conditional distribution must be understood as an asymptotic statistical interpretation of maximum likelihood, not as a literal description of the observed data. Teacher-forced training and free-running generation can also diverge, a phenomenon related to exposure bias \cite{he2021}.

\section{The Ergodicity Requirement}

For training on realized trajectories to inform the marginal conditional distribution $p_{\mathrm{marg}}$, the observed corpus must behave as a sufficiently representative sample from a stable marginalized language process. A simplified requirement is
\[
\widehat{p}_N(x_{t+1}\mid x_{\leq t})\longrightarrow p_{\mathrm{marg}}(x_{t+1}\mid x_{\leq t})
\]
as corpus size $N\to\infty$. Such convergence requires assumptions analogous to those used when empirical frequencies or maximum-likelihood estimates are interpreted as estimates of an underlying stochastic process: stationarity, ergodicity, representativeness, adequate support coverage, and stability \cite{birkhoff1931,doob1953,bishop2006,cover2006}.

Natural language violates these assumptions in many ways. Language changes across time, domain, genre, author, institution, platform, ideology, and communicative purpose. Moreover, for long contexts, exact repetition of $x_{\leq t}$ is rare or absent. The empirical conditional law for a long exact context is usually not observable.

Therefore, what the model learns in practice is not a table of conditional probabilities. It learns parameterized regularities that compress observed trajectories across related contexts. Without representativeness and ergodic stability, $p_\theta$ should be interpreted as a fitted distribution over a historical archive rather than as a stable law of language.

\section{Marginalization Is Not Enough}

Suppose, for the sake of argument, that the ergodicity problem is solved and that
\[
p_\theta(x_{t+1}\mid x_{\leq t})\approx p_{\mathrm{marg}}(x_{t+1}\mid x_{\leq t}).
\]
This is not enough to make next-token prediction useful. The marginal distribution may be a poor guide if omitted circumstances $Z_t$ remain important after conditioning on the text.

The full conditional law is
\[
p_{\mathrm{full}}(x_{t+1}\mid x_{\leq t},z_t).
\]
The marginal law is useful only if
\[
p_{\mathrm{full}}(x_{t+1}\mid x_{\leq t},z_t)\approx p_{\mathrm{marg}}(x_{t+1}\mid x_{\leq t})
\]
for the relevant range of $z_t$. Equivalently, one needs approximate conditional independence:
\[
X_{t+1}\perp Z_t\mid X_{\leq t}.
\]
In information-theoretic terms, one needs
\[
I(X_{t+1};Z_t\mid X_{\leq t})\approx 0.
\]
When this quantity is small, the hidden circumstances add little predictive information once the text is known. When it is large, fluent continuation is epistemically weak because the correct continuation depends on variables not contained in the prompt.

This is a sufficiency condition. The observed textual prefix $X_{\leq t}$ must be an approximately sufficient statistic for the latent circumstances $Z_t$ with respect to predicting $X_{t+1}$.

If this condition fails, then even a perfect estimate of $p_{\mathrm{marg}}$ may be of limited use. The model will generate plausible continuations averaged over unobserved situations, while the correct continuation in the actual situation depends on variables not contained in the text.

\section{Mixed Training Regimes and Local Islands of Sufficiency}

The preceding sections treated the language process as if it were homogeneous. Real training corpora are not homogeneous. A language model is trained on an indistinct mass of language containing many different regimes, genres, tasks, and communicative practices; this heterogeneity is already visible in large-scale web-trained models such as GPT-style systems \cite{radford2019,brown2020}.

A useful abstraction is to write the training distribution as a mixture:
\[
\mathcal{D}=\sum_{k=1}^{K}\pi_k\mathcal{D}_k,
\]
where each component $\mathcal{D}_k$ corresponds to a different linguistic or task regime, for example
\[
\begin{aligned}
\mathcal{D}_k\in\{&\text{programming},\text{mathematics},\text{fiction},\text{journalism},\text{law},\\
&\text{dialogue},\text{social media},\text{textbooks},\ldots\}.
\end{aligned}
\]
Each regime has its own latent circumstances $Z_t^{(k)}$ and its own full conditional process
\[
p_k(x_{t+1}\mid x_{\leq t},z_t^{(k)}).
\]
The sufficiency condition is therefore not global. It is local to a regime:
\[
I_k(X_{t+1};Z_t^{(k)}\mid X_{\leq t})\approx 0.
\]
This condition may hold approximately in local islands of language, such as programming, formal mathematics, standardized bureaucratic forms, boilerplate legal text, API documentation, textbook exercises, or highly conventional technical genres.

It will fail in many other regions, such as open-world factual claims, historical interpretation, medical diagnosis, political judgment, personal advice, strategic decision-making, moral or legal evaluation, and scientific discovery outside well-textualized evidence.

The model is therefore trained on a mixed corpus containing both sufficient-context and insufficient-context regimes. This has several implications. First, the training objective does not explicitly tell the model whether a given prompt belongs to a regime where the textual prefix is sufficient. The model learns continuation statistics over the mixture. Second, success in sufficiency islands can create a misleading impression of general competence. Third, mixed training encourages transfer of stylistic confidence across regimes. The user experiences a homogeneous interface, while the epistemic status of the output changes radically across domains.

\section{Mixture Identifiability and Local Conditional Laws}

For a model trained on heterogeneous data to learn the correct text-only conditional distribution, the past sequence must contain enough information to identify, or at least probabilistically infer, which regime generated it. If the corpus is generated by a mixture
\[
\mathcal{D}=\sum_{k=1}^{K}\pi_k\mathcal{D}_k,
\]
where each regime $k$ has a text-only conditional law
\[
p_k(x_{t+1}\mid x_{\leq t}).
\]
The global marginal text-only conditional is then not a single homogeneous law but a mixture conditional:
\[
p_{\mathrm{mix}}(x_{t+1}\mid x_{\leq t})
=\sum_{k=1}^{K}p(k\mid x_{\leq t})p_k(x_{t+1}\mid x_{\leq t}).
\]
Thus, in principle, a model can learn the correct mixture conditional only if it learns both $p(k\mid x_{\leq t})$, the posterior probability that the prefix belongs to regime $k$, and $p_k(x_{t+1}\mid x_{\leq t})$, the local continuation law inside that regime.

This is possible when the prefix carries enough information to classify the regime. 

Prefixes such as \texttt{def merge\_sort(arr):}, \texttt{begin\{proof\}}, or \texttt{SELECT customer\_id FROM} strongly identify programming, mathematical proof, and SQL-like regimes. 

In such cases,
\[
p(k^\ast\mid x_{\leq t})\approx 1,
\]
and therefore
\[
p_{\mathrm{mix}}(x_{t+1}\mid x_{\leq t})\approx p_{k^\ast}(x_{t+1}\mid x_{\leq t}).
\]
This is the favorable case. The model behaves locally as if it had selected a specialized conditional law.

By contrast, prefixes such as ``The answer is,'' ``The cause was,'' or ``The correct interpretation is'' may not identify a sufficiently precise regime or latent situation. Then $p(k\mid x_{\le t})$ remains diffuse, and the resulting conditional distribution is a blended average over heterogeneous possibilities.

This distinction explains familiar model behavior: prompt sensitivity, style-content confusion, false authority, local competence, and cross-regime contamination. The model may identify the genre correctly while failing to identify the epistemic situation. It may produce the style of a legal opinion without the evidence required for one, or the style of a scientific explanation without the data needed to justify it.

Accordingly, the learnability of a correct text-only conditional in heterogeneous training requires at least three assumptions:
\begin{enumerate}
\item the mixture weights and component distributions are sufficiently stable;
\item the corpus is representative and ergodic with respect to the mixture process;
\item the prefix permits reliable regime inference.
\end{enumerate}
Under these assumptions, a model may approximate
\[
p_\theta(x_{t+1}\mid x_{\leq t})\approx p_{\mathrm{mix}}(x_{t+1}\mid x_{\leq t}).
\]
But this is still only the text-only mixture conditional. It is not the full conditional process of language production.

Moreover, usefulness requires an additional local condition. For the inferred regime $k$, the omitted circumstances must be conditionally irrelevant:
\[
I_k(X_{t+1};Z_t^{(k)}\mid X_{\leq t})\approx 0.
\]
Thus, we obtain a two-stage criterion.

\begin{center}
\begin{tabular}{@{}p{0.25\linewidth}p{0.65\linewidth}@{}}
\toprule
Condition & Requirement \\
\midrule
Learnability & $p_\theta(x_{t+1}\mid x_{\leq t})\approx p_{\mathrm{mix}}(x_{t+1}\mid x_{\leq t})$ requires stable, ergodic, identifiable mixture structure. \\
Usefulness & $I_k(X_{t+1};Z_t^{(k)}\mid X_{\leq t})\approx 0$ requires local sufficiency inside the inferred regime. \\
\bottomrule
\end{tabular}
\end{center}

The two questions must not be confused. A model may learn the correct text-only mixture conditional and still generate continuations that are useless or misleading for the actual world situation.

\section{Known Phenomena in the Present Framework}

Many language models related phenomena can be understood as manifestations of a common statistical structure: the learned text-only marginal distribution is useful only under local sufficiency and adequate regime identification.
The most relevant of these are listed in the following table.

\begin{table}[H]
\centering
\begin{tabular}{p{0.34\textwidth}p{0.54\textwidth}}
\toprule
Known phenomenon & Interpretation in this framework \\
\midrule
Hallucination & Generation from a plausible text-only marginal when the actual continuation depends on omitted latent state \citet{ji2023,huang2023,maynez2020}. \\
Prompt sensitivity & Movement of posterior mass across latent mixture regimes \citet{min2022,xie2022}. \\
RAG improvement & Partial textualization of missing circumstances \citet{lewis2020,borgeaud2022}. \\
RAG failure & Retrieved material does not make residual latent state conditionally irrelevant. \\
Tool use & External access to facts, calculations, executions, or states outside the model-induced distribution \citet{karpas2022,schick2023,yao2023,gao2023}. \\
Temperature effects & Transformation of $p_{\theta}$ without restoration of omitted context \citet{holtzman2020}. \\
Model collapse & Recursive fitting to model-induced distributions rather than the target human marginal process \citet{shumailov2023,shumailov2024,alemohammad2024}. \\
Style-content confusion &
The model may identify the genre correctly while failing to identify the epistemic situation.  \\
False authority &
The output  has the surface form of explanation without the informational basis required for explanation.\\
Local competence &
The model may appear genuinely competent in domains where prefixes strongly identify the regime and contain most of the relevant state.\\
Cross-regime contamination &
The model may import patterns from nearby textual regimes that are statistically similar but epistemically inappropriate. This is especially likely when $p(k\mid x_{\leq t})$ is diffuse.\\
\bottomrule
\end{tabular}
\caption{Known LLM phenomena interpreted through the marginalization and local-sufficiency framework.}
\end{table}

\section{Programming as a Favorable Case}

Programming is a domain where both mixture identifiability and local sufficiency often hold approximately. More generally, modern language models display strong few-shot behavior in structured textual regimes, although such behavior should not be confused with guaranteed symbolic correctness \cite{brown2020,wei2022}. Programming prefixes frequently identify the regime with high probability. Code syntax, file structure, comments, imports, error messages, and programming-language-specific conventions make $p(k\mid x_{\leq t})$ sharply concentrated.

Moreover, much of the relevant latent state can be textualized. Suppose the context contains
\[
\begin{aligned}
X_{\leq t}=(&\text{natural language specification},\text{design notes},\text{imports},\\
&\text{previous code},\text{tests},\text{error messages},\text{documentation snippets}).
\end{aligned}
\]
In many programming tasks, the relevant latent state includes
\[
\begin{aligned}
Z_t=(&\text{intended function},\text{API constraints},\text{syntax rules},\\
&\text{type constraints},\text{test behavior},\text{runtime errors}).
\end{aligned}
\]
Much of this latent state can be inserted into the prefix. Thus, approximately,
\[
Z_t\approx g(X_{\leq t}),
\]
and therefore
\[
p_{\mathrm{full}}(x_{t+1}\mid x_{\leq t},z_t)\approx p_{\mathrm{full}}(x_{t+1}\mid x_{\leq t}).
\]
Programming is favorable for several reasons: syntax is explicit; many constraints are local and textual; prior code strongly constrains future code; specifications can often be written in text; error messages and tests externalize hidden state; correctness can often be checked by execution; and training corpora contain many repeated structural patterns.

Programming is not magically solved by next-token prediction. The more precise claim is that programming often satisfies
\[
I(X_{t+1};Z_t\mid X_{\le t})\ll I(X_{t+1};Z_t)
\]
more strongly than ordinary factual discourse. Correctness can also be externally checked through compilation, unit tests, type systems, execution, and formal methods. This distinguishes plausible code from executable or correct code.

 In summary: programming is a domain where the textual prefix often identifies the regime and contains much of the relevant state needed for continuation.

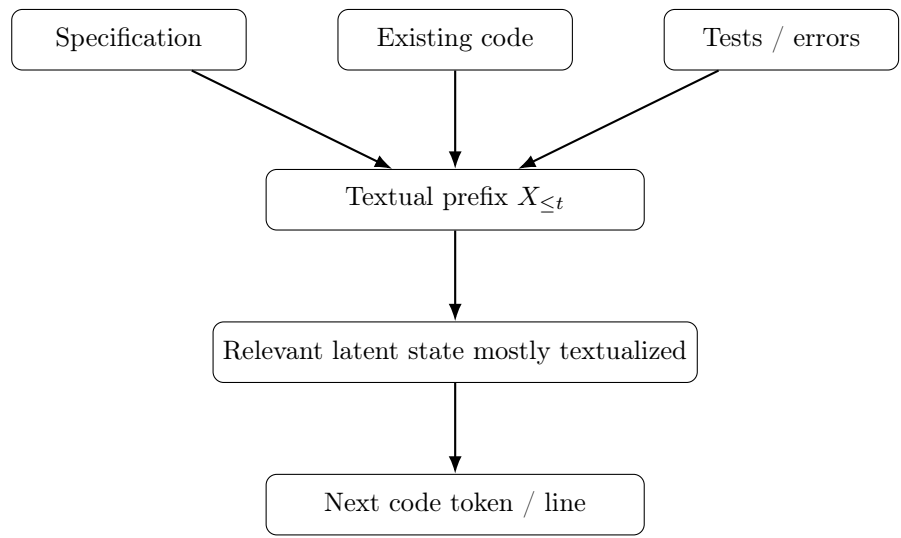
\begin{figure}[ht]
\centering
\begin{tikzpicture}[node distance=1.2cm, every node/.style={font=\small}, box/.style={draw,rounded corners,align=center,minimum width=3.1cm,minimum height=0.8cm}, arr/.style={-Latex,thick}]
\node[box] (spec) {Specification};
\node[box,right=of spec] (code) {Existing code};
\node[box,right=of code] (tests) {Tests / errors};
\node[box,below=1.3cm of code,minimum width=5cm] (prefix) {Textual prefix $X_{\leq t}$};
\node[box,below=of prefix,minimum width=5cm] (state) {Relevant latent state mostly textualized};
\node[box,below=of state,minimum width=5cm] (next) {Next code token / line};
\draw[arr] (spec) -- (prefix);
\draw[arr] (code) -- (prefix);
\draw[arr] (tests) -- (prefix);
\draw[arr] (prefix) -- (state);
\draw[arr] (state) -- (next);
\end{tikzpicture}
\caption{Programming as a favorable regime: specifications, previous code, tests, and errors often textualize the latent task state.}
\end{figure}

\section{RAG as Textualization of Latent Circumstances}

Retrieval-augmented generation can be understood as an attempt to improve sufficiency \citet{lewis2020,borgeaud2022}. Let $R_t$ denote retrieved material:
\[
R_t=(r_1,\ldots,r_k).
\]
The model now conditions on
\[
p_\theta(x_{t+1}\mid x_{\leq t},R_t).
\]
RAG is useful only if retrieved material reduces the remaining dependence on hidden circumstances:
\[
I(X_{t+1};Z_t\mid X_{\leq t},R_t)<I(X_{t+1};Z_t\mid X_{\leq t}).
\]
The strongest useful case is
\[
I(X_{t+1};Z_t\mid X_{\leq t},R_t)\approx 0.
\]
Equivalently,
\[
p_{\mathrm{full}}(x_{t+1}\mid x_{\leq t},z_t,R_t)\approx p_{\mathrm{full}}(x_{t+1}\mid x_{\leq t},R_t).
\]
Thus, RAG does not add knowledge symbolically to the model. It attempts to turn missing circumstances into additional text.

If the retrieved material does not capture the relevant omitted state, RAG merely adds more text. It may improve topicality, citation appearance, or stylistic confidence without solving the underlying epistemic problem. In a heterogeneous training setting, RAG must satisfy a regime-specific condition:
\[
I_k(X_{t+1};Z_t^{(k)}\mid X_{\leq t},R_t)\approx 0.
\]
This is especially important in educational RAG systems. A course RAG is useful only when the retrieved course material is sufficient for the student's question. If the missing context is the student's misconception, the intended level of the course, the exact theorem assumptions, or the teacher's notation, and the RAG material does not capture these variables, the model may still generate a plausible but misleading answer.

\section{Tool Use as External Access to Non-Textual State}

Tool use is stronger than RAG, but it is not automatically sufficient either. A calculator, theorem prover, compiler, database, search engine, or code interpreter can access or generate information not contained in the language model's predictive distribution. Tool use, skill-like modules, and instruction-following interfaces should therefore be treated as system-level orchestration rather than pure language modeling \cite{ouyang2022,karpas2022,schick2023,yao2023,gao2023}.

Let $A_t$ be the output of an external tool:
\[
A_t=\tau(X_{\leq t}).
\]
The model conditions on
\[
p_\theta(x_{t+1}\mid x_{\leq t},R_t,A_t).
\]
Tool use is useful only if the tool output reduces the remaining dependence on hidden circumstances:
\[
I(X_{t+1};Z_t\mid X_{\leq t},R_t,A_t)<I(X_{t+1};Z_t\mid X_{\leq t},R_t).
\]
The strongest case is
\[
I(X_{t+1};Z_t\mid X_{\leq t},R_t,A_t)\approx 0.
\]
Equivalently,
\[
p_{\mathrm{full}}(x_{t+1}\mid x_{\leq t},z_t,R_t,A_t)\approx p_{\mathrm{full}}(x_{t+1}\mid x_{\leq t},R_t,A_t).
\]
Thus, RAG and tools are not automatically proxies for missing context. They are proxies only insofar as conditioning on them makes the remaining latent circumstances irrelevant, or at least substantially less informative for the continuation.

This explains why tool-augmented systems can be more reliable than plain language models. They do not make the language model internally more truthful. They change the conditioning information available to generation and delegate parts of the task to systems with different objectives and guarantees.

\section{Consequences for Temperature}

Temperature modifies the model-induced distribution. Decoding choices are known to strongly affect generated text quality, diversity, and degeneration \cite{holtzman2020}:
\[
p_{\theta,T}(i\mid x_{\leq t})=
\frac{\exp(\ell_i/T)}{\sum_j\exp(\ell_j/T)}.
\]
But temperature does not solve the marginalization problem. It does not restore omitted circumstances. It does not make $p_{\theta,T}$ a sample from $p_{\mathrm{full}}$. Nor does it ensure that $p_{\theta,T}$ approximates $p_{\mathrm{marg}}$, unless the training process already produced such an approximation.

Temperature only changes how broadly the model samples from its own learned distribution. When the prefix is insufficient, higher temperature often increases diversity at the cost of epistemic reliability. When the prefix is highly sufficient, moderate stochasticity may be useful for exploring multiple valid continuations. This explains why sampling can be useful for creative writing or alternative code completions but dangerous for factual claims \citet{holtzman2020}.

\section{Synthetic Contamination and Recursive Training}

The preceding analysis also clarifies a further problem that becomes central once language-model outputs are injected into future training data. The problem is not merely that synthetic data may be low quality. It is not just a "regression to the mean" kind of problem we would have each time we use model fitted data added in a dataset. The deeper issue is that model-generated sequences are not samples from the full human language process, and need not even be samples from the marginalized text-only law discussed above. They are samples from a model-induced distribution, possibly after decoding transformations such as temperature, nucleus sampling, filtering, alignment, or tool-mediated post-processing.

Let \(P\) denote the target marginalized human text process and let \(Q_{\theta,T}\) denote the distribution of sequences produced by a model with parameters \(\theta\) and decoding rule \(T\). If generated data are mixed into the next training corpus, the effective training distribution becomes schematically
\[
P_{\alpha}=(1-\alpha)P+\alpha Q_{\theta,T}, \qquad 0\leq \alpha\leq 1.
\]
The next model trained on this corpus is then pushed toward \(P_{\alpha}\), not toward \(P\). If \(Q_{\theta,T}=P\), no distortion is introduced. But this equality is precisely what the previous sections show cannot be assumed. It would require the original model to have learned the correct marginalized distribution under strong ergodicity, representativeness, mixture-identifiability, and local sufficiency conditions. If these requirements fail, generated data introduce a biased component into the empirical training process.

The point can be made recursively. Suppose models are trained through generations according to
\[
P_{n+1}=(1-\alpha)P+\alpha Q_{\theta_n,T_n},
\]
where \(Q_{\theta_n,T_n}\) is the distribution generated by the previous model. Unless \(Q_{\theta_n,T_n}\) is already a faithful sample from the target process, approximation errors, missing tails, regime confusions, and hallucinated structures can be reintroduced as training evidence. The training procedure then begins to fit not only human language trajectories but also distortions created by previous models.

This connects the present argument with the literature on model collapse and self-consuming generative models. Shumailov et al. describe model collapse as arising when models recursively train on generated data and write that ``model collapse is universal among generative models that recursively train on data generated by previous generations'' \cite{shumailov2024}. Their earlier formulation emphasizes that models can ``forget the true underlying data distribution'' and that ``tails of the original content distribution disappear'' \cite{shumailov2023}. Alemohammad et al. analyze related autophagous loops and conclude that, without enough fresh real data, future models see quality or diversity progressively decrease \cite{alemohammad2024}.

The framework developed in this paper gives a complementary interpretation of these results. If a model-generated sequence comes from a regime in which the prefix was not sufficient for the omitted context, then the sequence may encode a plausible continuation rather than a sample from the true marginalized conditional process. If the model failed to identify the correct local regime, the generated sample may mix incompatible regimes. If RAG or tool outputs were used but did not satisfy the conditional sufficiency requirement, then the generated sequence may contain the surface form of grounded language without the relevant grounding state. If no truth-checking mechanism is applied, false statements may be injected as ordinary language evidence.

Consequently, AI-generated data are not neutral additions to the corpus. They are observations from a derived distribution whose relation to the target language process depends on precisely the assumptions analyzed above. If the generating models do not satisfy those assumptions, adding their outputs to future training data makes the possibility of estimating the human marginalized conditional distribution even weaker. In mixture notation, synthetic contamination adds components
\[
\mathcal{D}_{\mathrm{AI},j}
\]
whose conditional laws are not human regime conditionals \(p_k\), but model-induced approximations \(q_j\). The effective mixture becomes
\[
\mathcal{D}'=(1-\alpha)\sum_k \pi_k\mathcal{D}_k+
\alpha\sum_j \rho_j\mathcal{D}_{\mathrm{AI},j}.
\]
A model trained on \(\mathcal{D}'\) learns the conditional law of this contaminated mixture, not the original human-language mixture. If the synthetic components underrepresent rare regimes, omit distributional tails, overproduce high-probability stylistic patterns, or contain hallucinated facts, these distortions become part of the next training target.

This also interacts with the sufficiency criterion. For a generated sequence to be a safe training sample, it is not enough that it be fluent. One would need, at minimum, that the model which generated it operated in a regime satisfying
\[
I_k(X_{t+1};Z_t^{(k)}\mid X_{\leq t},R_t,A_t)\approx 0,
\]
and that the resulting generated statement be valid or truth-checked when the domain requires truth rather than mere continuation. Otherwise, synthetic data may amplify exactly the regimes in which next-token prediction is least epistemically reliable.

Thus, the present paper implies a stricter view of recursive training than the usual warning that models should not be trained on ``bad'' synthetic data. The issue is structural. Model-generated text is trustworthy as training evidence only when the generating process itself satisfied the learnability, identifiability, sufficiency, and verification requirements relevant to the domain. Without these conditions, synthetic contamination makes the original ergodic-estimation problem harder rather than easier.

\section{Educational Implications}

Recent work has suggested LLM-based and retrieval-augmented systems as scalable components of personal or adaptive tutors, continuing a long tradition of research on one-to-one tutoring and intelligent tutoring systems \cite{bloom1984,graesser2005,vanlehn2011,ma2014}. More recent educational discussions of LLMs emphasize their possible use for feedback, explanation, exercise generation, and tutoring, while also stressing risks of hallucination, bias, privacy loss, and over-reliance \cite{kasneci2023}. In this context, RAG has been proposed as a way to build course-specific tutors by grounding responses in curated instructional material rather than relying only on the parametric distribution of a general language model \cite{lewis2020,borgeaud2022,dong2023}.

The analysis developed above can therefore be read as a set of conditions under which such educational uses may actually be useful. A RAG-based tutor is not reliable merely because it retrieves course notes, nor because it answers in a pedagogically fluent style. It is useful when the educational task belongs to a regime that the prompt and retrieved material identify with sufficient precision, and when the augmented context is close to sufficient for the latent pedagogical state relevant to the student's question.

In the notation of the paper, the educational usefulness condition is local to the task regime:
\[
I_k(X_{t+1};Z_t^{(k)}\mid X_{\leq t},R_t,A_t)\approx 0.
\]
Here \(Z_t^{(k)}\) includes not only the objective subject matter, but also educationally relevant latent variables such as the student's misconception, the intended level of the course, the notation adopted by the teacher, the exact assumptions of the theorem or exercise, and the learning goal of the interaction. The retrieved material \(R_t\) and any tool output \(A_t\) are useful only insofar as they reduce the residual dependence on these variables.

Thus, educational language-model systems are most justified when:
\begin{itemize}
    \item the relevant task state can be textualized;
    \item the course materials retrieved by the RAG system are actually sufficient for the student's question;
    \item the local regime is well identified, as in a specific course, exercise type, programming assignment, or mathematical procedure;
    \item correctness can be externally checked, for example through tests, calculations, symbolic systems, or teacher review;
    \item the student is asked to inspect, criticize, and compare the output rather than merely accept it.
\end{itemize}

They are least reliable when the correct answer depends on hidden facts, the prompt under-specifies the situation, the retrieval system fails to capture the relevant course context, the student's misconception is not represented in the prompt, or no external verification is available.

Programming tutors, mathematical exercise generators with symbolic checking, and RAG systems based on carefully curated course materials therefore have a stronger theoretical justification than open-ended factual tutors without retrieval or verification. This does not mean that their outputs are guaranteed to be true or pedagogically appropriate. The present section only states conditions under which the language-model distribution may become educationally useful. It does not solve the separate truth-checking and validity-checking problem discussed in the introduction: even when the conditioning context is sufficient for plausible continuation, correctness still requires either domain conditions under which likely continuations are usually true or external mechanisms of verification.

\section{Prompts, Non-Observed Conditionals, and the Limits of Context Injection}

The preceding sections considered the conditions under which a language model trained on observed token trajectories may approximate a marginal text-only conditional distribution. We now consider a further limitation: even when the prompt contains full information about the relevant context, the model cannot reliably follow a conditional distribution that does not correspond to trajectories observed, directly or indirectly, in training.

Let the full language process be

\[
p_{\mathrm{full}}(x_{t+1}\mid x_{\leq t},z_t),
\]

where \(z_t\) denotes the relevant non-textual circumstances. Suppose that, at inference time, the prompt is enriched with a textual representation \(c_t\) of those circumstances. The model is then queried on an extended prefix

\[
\tilde{x}_{\leq t}=(x_{\leq t},c_t).
\]

A common intuition is that, if \(c_t\) contains all relevant information, then the model should be able to condition correctly on it. This intuition is too strong. The model can exploit \(c_t\) only if training has exposed it to sufficiently many trajectories in which similar contextual material played the relevant conditioning role. In other words, the model must have learned a conditional law of the form

\[
p(x_{t+1}\mid x_{\leq t},c_t).
\]

If the relevant conditional pattern was not observed in the training corpus, then the prompt does not create it. The prompt changes the input trajectory, but it does not supply missing conditional data to the training process.

This can be interpreted as a support or ergodicity failure. For the model to learn the conditional distribution associated with the extended context, the training process must contain enough representative occurrences of the relevant kind of conditioning event. Formally, one would need convergence of the form

\[
\widehat{p}_N(x_{t+1}\mid x_{\leq t},c_t)
\longrightarrow
p_{\mathrm{marg}}(x_{t+1}\mid x_{\leq t},c_t).
\]

If the event class \((x_{\leq t},c_t)\) is absent, extremely rare, or not part of an ergodic component of the training distribution, this convergence cannot occur. The model then has no learned conditional distribution corresponding to the requested situation. It may still generate a fluent continuation by analogy with nearby trajectories, but such a continuation is not an estimate of the desired conditional law.

This makes precise the sense in which a prompt is not ``new information for the model.'' It is new text placed in the conditioning sequence. It may redirect generation toward regions of the learned distribution, but it cannot substitute for missing training experience. A prompt can select among learned conditional behaviors; it cannot by itself install a conditional behavior that was not learned.

The same point applies to RAG and tool outputs. Retrieved material \(R_t\) and tool output \(A_t\) enlarge the conditioning sequence:

\[
p_\theta(x_{t+1}\mid x_{\leq t},R_t,A_t).
\]

However, the model can exploit this enlarged context only if its training has included sufficiently similar situations in which such added material correctly constrained continuation. Otherwise, RAG or tool output may be present in the prompt but not function as an effective conditioning variable. It may improve topical relevance or surface coherence without inducing the desired conditional behavior.

Thus, two conditions must be distinguished. The first is informational sufficiency:

\[
I(X_{t+1};Z_t\mid X_{\leq t},R_t,A_t)\approx 0.
\]

This says that the added material contains enough information to make the hidden context irrelevant. The second is learned conditional availability: the model must have been trained on enough trajectories for which similar augmented prefixes were associated with the appropriate continuations. Without the second condition, even a fully informative prompt may fail.

This is especially clear in formal domains. Suppose a prompt fully specifies an arithmetic operation or a new formal rule. If the model has not learned the relevant conditional structure from training, it is not guaranteed to follow the rule. The correctness of the output depends not on the presence of the rule in the prompt alone, but on whether the model has learned how such rules constrain continuations. This is why the success of language modeling and the truth or validity of generated statements remain distinct properties. A model may fit language trajectories well and still lack a mechanism for checking whether a generated continuation is true, valid, or rule-compliant.

This observation also clarifies the role of in-context learning. In-context examples can guide a model when they activate conditional patterns already learned during training \cite{brown2020, min2022, xie2022}. They do not guarantee arbitrary rule acquisition from the prompt. The prompt may induce local adaptation in generation, but this adaptation remains constrained by the model-induced distribution. In this sense, prompting is better understood as conditional selection within learned trajectories than as the direct insertion of new knowledge or new rules.

The general implication is the following. Even if the prompt contains full information about the relevant context, the model can only use that information reliably when the corresponding conditional relation belongs to the learned support of the training distribution. Otherwise the problem is, in the terminology of the present paper, another failure of ergodicity or support coverage. The relevant conditional distribution was not observed, and therefore cannot be recovered merely by writing the conditioning variables into the prompt.

\section{Conclusion}

The usefulness of next-token prediction depends on three separate conditions.

First, the model-induced distribution can be interpreted as an approximation to a marginal text-only conditional law only under strong assumptions of representativeness, stationarity, and ergodicity of the observed corpus.

Second, in a heterogeneous corpus, the prefix must allow the model to infer the relevant local regime:
\[
p(k^\ast\mid X_{\leq t})\approx 1
\]
or at least to estimate the appropriate mixture conditional:
\[
p_{\mathrm{mix}}(x_{t+1}\mid x_{\leq t})=
\sum_k p(k\mid x_{\leq t})p_k(x_{t+1}\mid x_{\leq t}).
\]
Third, even if this approximation succeeds, the resulting text-only law is useful only when the observed prefix is an approximately sufficient statistic for the latent circumstances relevant to continuation.

The central criterion is
\[
I(X_{t+1};Z_t\mid X_{\leq t})\approx 0.
\]
In heterogeneous corpora, this criterion becomes regime-specific:
\[
I_k(X_{t+1};Z_t^{(k)}\mid X_{\leq t})\approx 0.
\]
Some parts of language form local islands where the condition approximately holds. Programming is a favorable example because specifications, code, tests, error messages, and documentation often textualize much of the relevant latent state.

Many other parts of language do not satisfy the condition. In such regions, the model produces plausible continuation rather than situation-specific truth.

RAG and tool use can be understood as attempts to improve sufficiency by adding retrieved evidence or externally computed information. But they are useful as proxies for missing context only when
\[
I_k(X_{t+1};Z_t^{(k)}\mid X_{\leq t},R_t,A_t)\approx 0.
\]

Finally, even a prompt containing the full relevant context cannot make the model follow a conditional law that was not represented in the training trajectories. Prompting, RAG, and tool outputs can only condition generation through structures already made available by the learned distribution; otherwise the failure is again one of support, representativeness, and ergodicity.

Language model usefulness and reliability depends on the validity of these conditions. Their success depends not only on model size, architecture, decoding strategy, or corpus scale, but on whether the relevant world has been adequately turned into text and whether the local regime has been adequately identified. 

Even then, distributional success is not the same as truth. The additional question - whether generated statements are factual, valid, executable, or mathematically correct - requires either domain conditions under which likely continuations are usually true or external checking mechanisms such as retrieval, execution, formal verification, or other tools \cite{ji2023,huang2023,gao2023,schick2023}.

\newpage
\bibliographystyle{plainnat}

\end{document}